\pdfoutput=1

\documentclass[11pt]{article}

\usepackage{EACL2023}

\usepackage{times}
\usepackage{latexsym}
\usepackage{xcolor}
\usepackage{makecell}
\usepackage{times}
\usepackage{latexsym}
\usepackage{hyperref}
\usepackage{inconsolata}
\usepackage{url}
\usepackage{makecell}
\usepackage{multirow}
\usepackage{graphicx}
\usepackage{caption}
\usepackage{subcaption}
\usepackage{amsmath}
\usepackage{bbm}
\usepackage{booktabs}
\usepackage{algorithmic}
\usepackage{amssymb}
\usepackage{bm}
\usepackage[ruled,vlined,linesnumbered]{algorithm2e}
\usepackage{enumitem}
\usepackage{pifont}
\usepackage{makecell}
\usepackage{pbox}

\usepackage{color}

\usepackage[T1]{fontenc}

\usepackage[utf8]{inputenc}

\usepackage{microtype}

\usepackage{inconsolata}

\title{COVID-VTS: Fact Extraction and Verification on Short Video Platforms}

\author{Fuxiao Liu \And Yaser Yacoob \\ \\ University of Maryland, College Park\\ \{fl3es, yaser, abhinav\}@umd.edu \And Abhinav Shrivastava \\}

\begin{document}
\maketitle
\begin{abstract}
We introduce a new benchmark, \texttt{COVID-VTS}, for fact-checking multi-modal information involving short-duration videos with COVID19-focused information from both the real world and machine generation. We propose, \texttt{TwtrDetective}, an effective model incorporating cross-media consistency checking to detect token-level malicious tampering in different modalities, and generate explanations. Due to the scarcity of training data, we also develop an efficient and scalable approach to automatically generate misleading video posts by event manipulation or adversarial matching. We investigate several state-of-the-art models and demonstrate the superiority of \texttt{TwtrDetective}.
\end{abstract}

\section{Introduction}
\begin{figure*}[t]
    \centering
      \includegraphics[width=1\textwidth]{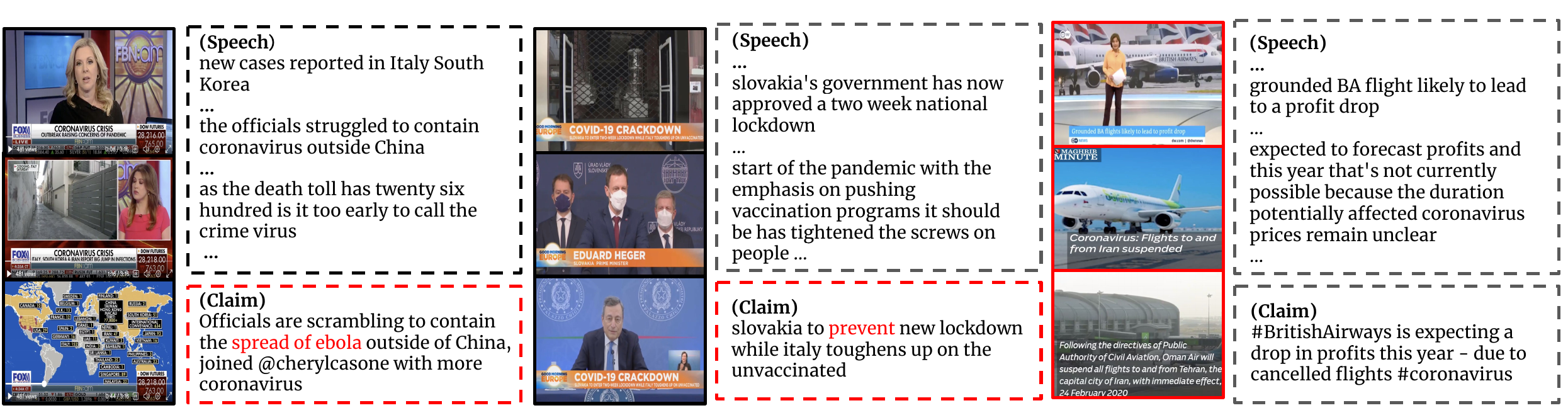}
    \caption{Examples of different inconsistent taxonomies from \texttt{COVID-VTS}, which are generated by our automatic manipulation tool. The red box indicates the inconsistent modality. Our task is to independently decide whether the video-text pair is consistent and point out which modality is fake. (1) The event argument in the claim is modified; (2) The event trigger in the claim is tampered; (3) The short video is curated by adversarial matching.}
    \vspace{-0.1in}
    \label{fig:Automatic_tool}
\end{figure*}

The proliferation of misinformation in social media poses a serious threat to our society, especially during the COVID-19 pandemic. Therefore, it is necessary to develop automatic fact-checking tools to verify the claims propagated online.

Recent fact-checking work \cite{thorne2018fever, wadden2020fact, augenstein2019multifc,fung2021infosurgeon, bekoulis2021review} investigated automatic misinformation detection by developing various benchmark datasets as well as state-of-the-art neural network architectures involving sources such as Wikipedia pages, tables, news articles, and scientific articles. However, due to the lack of publicly available benchmarks, fact checking is still challenging on short video platforms, such as TikTok, Twitter, and Instagram.

Verifying the factual correctness of claims in short video platforms poses new challenges. Firstly, multi-modal misinformation that leads to short video posts is more misleading than using just textual content \cite{micallef2022cross}, since the claims usually only tamper with subtle elements of the factual information from the source video. Furthermore, short videos \cite{shang2021multimodal} containing diverse scene shifts, human activities, and cross-modal information, greatly increase the complexity and ambiguity of the video content. Moreover, current methods explored the authenticity of GAN-generated video with unimodal analysis \cite{guera2018deepfake}. Nevertheless, they cannot be directly applied to the short video platform, where the video content is often intentionally manipulated.

To tackle these challenges, we introduce \texttt{COVID-VTS}, a new benchmark dataset with trustworthy claims and corresponding good-quality videos, collected from Twitter video posts. \texttt{COVID-VTS} contains different inconsistent taxonomies and the inconsistency comes from different modalities. Examples are shown in Figure~\ref{fig:Automatic_tool}. We additionally introduce a novel approach to generate fake video posts automatically by manipulating event elements or adversarial matching \cite{luo2021newsclippings}. The advantage of our event manipulation tool is that it's able to control the polarization of semantics and track the manipulation object by editing a small component of the factual information. We apply quality control to delete unqualified generations and address the linguistic bias.

We propose \texttt{TwtrDetective}, a multimodal fact-checking model, where the input consists of a claim with the paired video, and the goal is to predict the consistency. \texttt{TwtrDetective} takes advantage of the \textit{Event Alert} module to precisely extract fine-grained factual details from the claim, as well as \textit{Pairwise Consistency Aggregation} module to effectively measure the consistency between each modality. \texttt{TwtrDetective} is also able to point out the inconsistent modality (e.g., see Figure~\ref{fig:model}). 

Experimental results show that \texttt{TwtrDetective} achieves higher detection accuracy over baselines on two datasets. Our main contributions are summarized as:
\begin{itemize}[nosep,leftmargin=*]
    \item We introduce \texttt{COVID-VTS}, a fact-checking dataset for short video platforms, consisting of 10k video text pairs with diverse scenes, more accessible modalities, and trustworthy claims from both the real world and machine generation.  
    \item We propose an effective approach to automatically generate large-scale verifiable, trustworthy as well as misleading video posts rather than employing human annotators.
    \item We propose \texttt{TwtrDetective}, a new explainable fact-checking framework for the short video platform, showing superior results on two challenging datasets with respect to baselines.
\end{itemize}
Our code is publicly available at \url{https://github.com/FuxiaoLiu/Twitter-Video-dataset}.

\label{sec:intro}

\section{Related Work}
The spread of misinformation has led to a growing body of research in automatic fact-checking. Many large scale datasets collected from Wikipedia and fact-checking websites were introduced, including FEVER \cite{thorne2018fever}, SciFact \cite{wadden2020fact}, UKP Snopes \cite{hanselowski2019richly}, MultiFC \cite{augenstein2019multifc}. However, the fake claims in these datasets are manually generated by humans, making it expensive to deploy at scale. Recently, some synthetic datasets were proposed to address this limitation. \cite{jiang2020hover} generated complex fake claims using word substitutions. \cite{saakyan2021covid} took advantage of the token-infilling ability from Masked Language Model to replace the salient tokens. Also, \cite{fung2021infosurgeon} formulated a novel data synthesis method by manipulating knowledge elements within the multi-modal knowledge graph. In order to alleviate the linguistics bias within the machine-generated claims, \cite{luo2021newsclippings} constructed the out-of-context captions by retrieving the real-world sentence with the similar semantics. In a similar vein our dataset, \cite{liu2020violin} constructed VIOLIN for the Video-Language Inference while all the statements are written by experts. \cite{wang2022misinformation} collected social media video posts from Twitter but the fake claims are constructed by random swap. In comparison, \texttt{COVID-VTS} is the first COVID-19 fact-checking dataset for short video platforms, containing rich information with diverse scenes, more accessible modalities as well as misleading claims from both the real world and machine generation.

Traditional fact-checking approaches \cite{zellers2019defending,schuster2020limitations, atanasova2019automatic} are mainly based on text. They fall short if the evidence stems from visual information. Recent multi-modal models \cite{fung2021infosurgeon,tan2020detecting,shang2021multimodal} are equipped with the ability to check consistency according to the information conveyed across modalities. In contrast, we propose a fact-checking model for the short video platforms with multimedia explanations that achieves higher accuracy to detect misinformation.

\label{sec:related_work}

\section{COVID-VTS Dataset Construction}
\texttt{COVID-VTS} contains 10k well-formed claims with the paired videos to support or refute the claims. In this section, we first describe the procedure to select the trustworthy and consistent video/claims pairs from Twitter video posts. Second, we also present our approach to automatically construct well-formed inconsistent video-claim pairs.

\subsection{Filtering for Authentic and Consistent Video Posts}
To assemble the \texttt{COVID-VTS} dataset, we used the Twitter scraper to collect over 100k English video posts using COVID related keywords and hashtags ranging from the end 2019.10 to 2022.8. Examples are shown in Table~\ref{fig:hashtags}. We retain one post for a video link to reduce the potential bias. To retrieve the speech from videos, we leverage the Speech-to-Text tool from IBM Cloud \cite{pitrelli2006ibm}. In addition, we leverage SimpleOCR \cite{ko2004simple} to recognize text on screen. After the preliminary filtering, we have several steps to select the authentic and consistent video posts:

\noindent\textbf{VERIFIED User Account.} According to the requirements from Twitter, only the authentic, notable, and active accounts will receive the blue verified badge (including official government, official company, etc.). Accounts that routinely post content that harasses, shames, or engaged in severe violations of our platform manipulation and spam policy are ineligible for blue badge. We collect the authentic posts from the verified accounts into \texttt{COVID-VTS} to improve the data quality.

\noindent\textbf{Claims Must be VERIFIABLE.} Event structures are essential to reveal the factual information of a sentence, since the overall semantics might be opposite if event elements are changed. Moreover, claims in \texttt{COVID-VTS} are supposed to be verifiable propositions whose truthfulness is determined by multi-modal evidences from the paired videos. Without the event structures, its truthfulness isn't verifiable. Thus, we delete instances with personal opinions or emotions. For example, \textit{'So proud of my boys! \#GetVaccinated'}, which has no factual information to be verified.

In practice, we first remove the claims without event structures, by using DYGIE++ \cite{wadden2019entity}, a state-of-art event extraction framework, pretrained on MECHANIC dataset. The event structures generated by DYGIE++ include event triggers and event arguments. For example in \textit{'Officials are scrambling to contain outbreaks of the coronavirus outside of China'}, \textit{'contain'} is the trigger which express the occurrence of the event, \textit{'officials'} and \textit{'outbreaks of the coronavirus'} are event arguments which play different roles in this event.

After selecting these verifiable claims, we manually check whether they are consistent with the associated videos. We also select verifiable and consistent video text pairs from the unverified accounts in order to increase the size of our dataset. Our hypothesis is that the video posts are trustworthy after the ``cleaning'' steps. Given this hypothesis, the main task is to analyze the inconsistency between different modalities. 

\begin{table}[t]
\small
\centering
	\centering
	\begin{tabular}{c}
		 \toprule       
	        \textbf{Keywords/Hashtags} \\ 
            \midrule
            \midrule  
         \pbox{7cm}{\textit{covid, \#covid, \#covid19, corona, \#coronavirus, maskup, pandemic, ICU, vaccine, \#vaccine, coronavirus, quarantine, \#quarantine, moderna, \#covidvariant, omicron, booster, mRNA, phizer, \#phizer, \#who, \#workfromhome, \#vaccinepassports, \#travelbans, \#social\_distance, n95}}\\
		\bottomrule
	\end{tabular}
 \caption{Summary of the covid related keywords and hashtags in the data collection process.}
    \label{fig:hashtags}
\end{table}

\subsection{Automatic \textsc{Inconsistent} Video-Claim Pairs Generation}
In this section, we describe the steps to automatically generate inconsistent video-claim pairs. Our dataset contains the following three inconsistent taxonomies: (1) Real video, Real speech and Inconsistent claim; (2) Real video, Inconsistent speech and real claim; (3) Inconsistent video, Real speech and Real claim.

\begin{table}[t]
\small
\centering
	\centering
	\begin{tabular}{cc}
		 \toprule       
	        &  \textbf{Most frequent Tokens} \\ 
            \midrule
            \midrule
       
		  \pbox{4cm}{\makecell[c]{Event Trigger\\ (Original Claims)}} &  
         \pbox{4cm}{\textit{fight, infect, protect, prevent, help, stop, quarantine, contain, confirm, threat, plunge, mandate, deal, cause}}\\
            \midrule
        \pbox{4cm}{\makecell[c]{Event Argument\\ (Original Claims)}} &  
         \pbox{4cm}{\textit{coronavirus, omicron, pfizer, covid, delta, moderna, booster, mask, lockdown, protest, social distance, ban, vaccine }}\\
            \midrule
        \pbox{4cm}{\makecell[c]{Event Trigger\\ (Generated Claims)}} &  
         \pbox{4cm}{\textit{produce, increase, create, develop, remove, protect, avoid, identify, support, generate, rule, allow, establish}}\\
            \midrule
        \pbox{4cm}{\makecell[c]{Event Argument\\ (Generated Claims)}} &  
         \pbox{4cm}{\textit{omicron, variant, death, cancer, hospital, WHO, community, medical service, drug, viruses, migration, media, ICU}}\\
  
		\bottomrule
	\end{tabular}
 \caption{Most frequent event triggers and arguments in original claims and generated claims.}
    \label{fig:frequent_keywords}
\end{table}

\begin{table*}[t]
\small
\centering
	\centering
	\begin{tabular}{cc}
		 \toprule        
	      \textbf{Original \textsc{\underline{True}} Claims}  &  \textbf{Generated \textsc{\underline{Fake}} Claims} \\ 
            \midrule
            \midrule
       
		  \pbox{7.5cm}{\textit{Officials are scrambling to contain \textcolor{blue}{outbreaks of the coronavirus} outside of China.}} &  \pbox{7.5cm}{\textit{Officials are scrambling to contain \textcolor{red}{the spread of ebola} outside of China.}}\\
            \midrule
    
            \pbox{7.5cm}{\textit{Moderna chairman \textcolor{blue}{getting vaccinated booster shots} is the only way to stop the virus.}} &  \pbox{7.5cm}{\textit{Moderna chairman \textcolor{red}{getting infected} is the only way to stop the virus.}}\\
            \midrule

        \pbox{7.5cm}{\textit{Fed chair powell warns omicron variant could \textcolor{blue}{dent} economic recovery.}} &  
        \pbox{7.5cm}{\textit{Fed chair powell warns omicron variant could \textcolor{red}{facilitate} economic recovery.}}\\\midrule

        \pbox{7.5cm}{\textit{Australians caught up in china's crisis have \textcolor{blue}{finally returned home} after 14 days quarantined on christmas island.}} &  
        \pbox{7.5cm}{\textit{Australians caught up in china's crisis have \textcolor{red}{lost home} after 14 days quarantined on christmas island.}}\\
  
		\bottomrule
	\end{tabular}
 \caption{A detailed look into the examples generated by our efficient automatic approach. The first two examples are the event-argument manipulation and the final two are the event-trigger manipulation.}
    \label{fig:example}
\end{table*}

\noindent\textbf{Inconsistent Claim Generation.} Inspired by \cite{liu2020violin}, we automatically generate fake-claims by controlling the polarization of semantics and tracking the manipulation object by modifying a small portion of the factual information in true claims. In this case, most of the statement remains true to the video content. This strategy can also alleviate the linguistic bias, which was introduced by \cite{fung2021infosurgeon, zellers2019defending}. In their datasets, the model correctly detects the fake claims without comparing different modalities. This is because all the fake claims are generated by language models.  Another advantage is that we can generate the large-scale training set automatically rather than employing human annotators.

Only event triggers or event arguments will be selected as the manipulated tokens \texttt{[mask]} depending on different intentions. We follow the procedures from \cite{nguyen2020bertweet} and use BERTWeet, pretrained on the Tweets related to the COVID-19 pandemic, to predict the domain-aware alternates of the event elements which can be subject to masking. In order not to introduce additional noise, we delete the candidates if their substituted tokens are not in the vocabulary of the original dataset to reduce the potential bias. Then, we feed the candidates and their corresponding true claims as input to the CrossEncoder model trained on Mutli-NLI dataset \cite{williams2017broad} and select the candidates with the \textsc{contradiction} label. If the candidate pair has the highest contradiction score, it will be assigned as the fake claim. Table~\ref{fig:example} presents examples generated by our efficient automatic approach. The first two examples are the event-argument manipulation and the final two are the event-trigger manipulation.

\begin{figure}[t]
    \centering
      \includegraphics[width=0.45\textwidth]{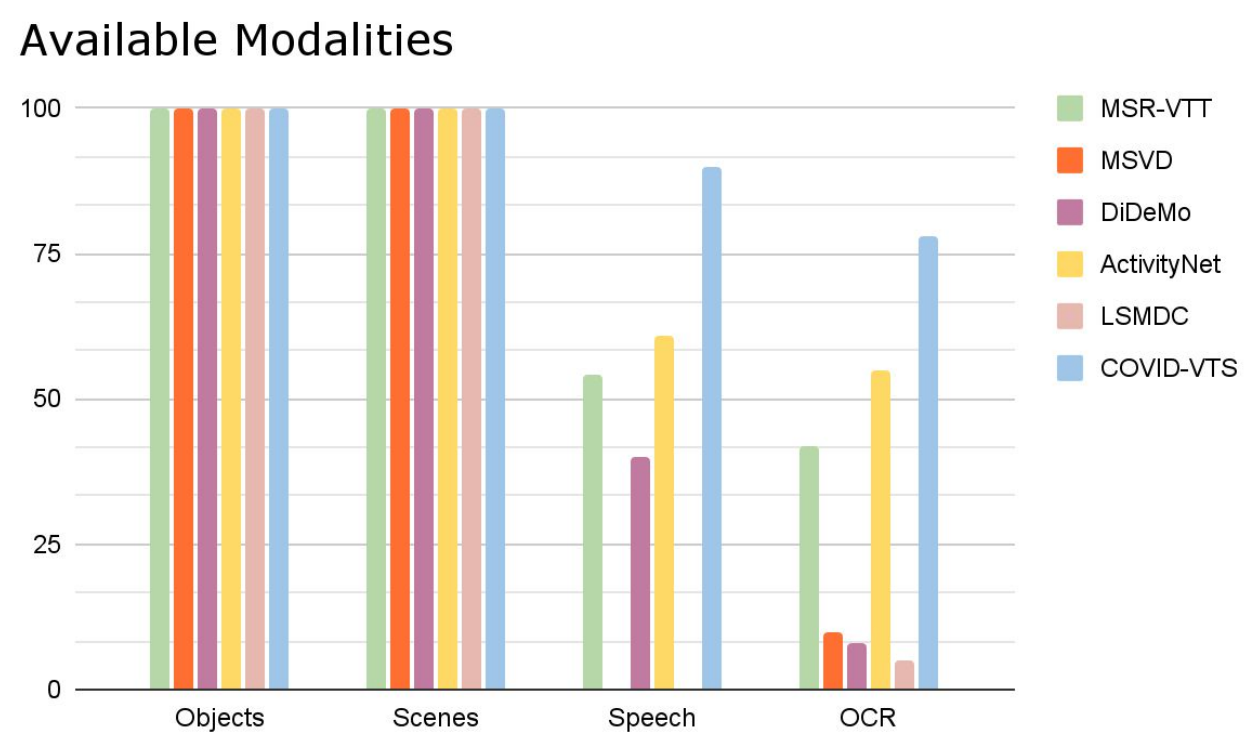}
    \vspace{-5pt}
    \caption{The histogram indicates COVID-VTS
is more accessible to speech content and text on the screen than previous datasets.}
    \vspace{-0.2in}
    \label{fig:available}
\end{figure}

Table~\ref{fig:frequent_keywords} shows the most frequent event elements in original claims and generated claims. In practice, we found that these frequent alternatives bring the additional linguistic bias. Models can learn to classify the claims as fake ones simply by detecting these frequent tokens without absorbing information from other modalities. Therefore, we alleviate this bias  by only keeping one claim for each alternative. Then replace unqualified claims by selecting the most similar claims from the dataset. Inconsistent pairs constructed this way focus more on the global understanding of the pairs. This rigorous setting makes the claims more challenging to distinguish by the analysis model, and in-depth reasoning is required to identify the fake content. 
\begin{table}[t]
\setlength\tabcolsep{4.3pt}
\centering
\small
\begin{tabular}{lcc}
\toprule
 & COVID-VTS & VIOLIN\\
\midrule
Average Caption Length      & 19.2 & 18.0\\
Average Speech Length         & 69.2 & 76.4\\
Average Video Length         & 26.5s & 35.2s \\
Named Entities (Sentence)         & 90.8\% & 10.3\% \\
Source         & Twitter & TV show \\
\bottomrule
\end{tabular}
\caption{
Summary of \texttt{COVID-VTS} dataset.
}
\vspace{-0.1in}
\label{tab:comparison_with_VIOLIN}
\end{table}

\begin{figure*}[t!]
    \centering
      \includegraphics[width=0.9\textwidth]{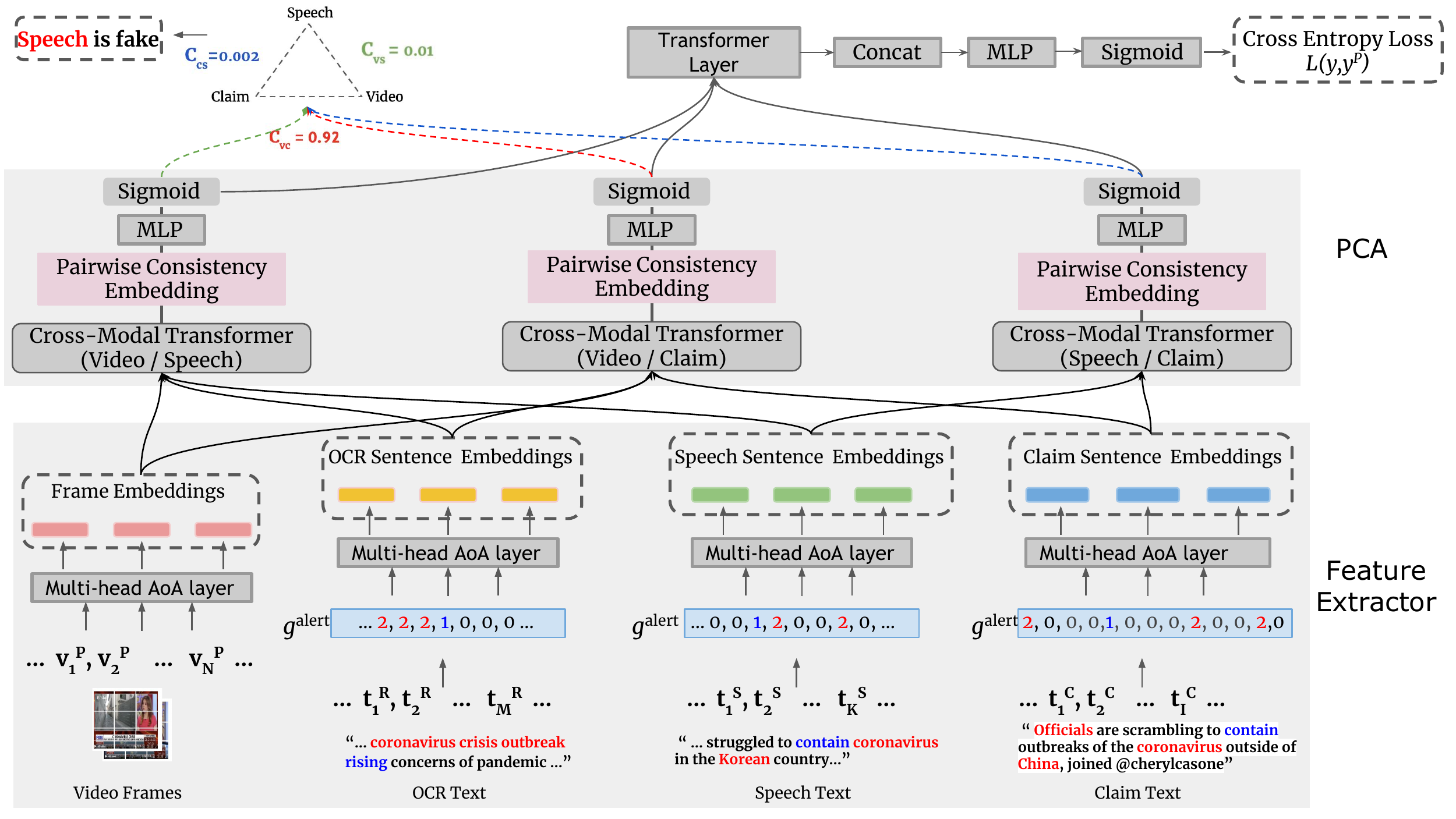}
    \caption{Overview of \texttt{TwtrDetective}. It detects the cross-modal inconsistency by comparing with video appearance, speech content, screen text and claims. $g^{alert}$ means the event alert module. In this example, \texttt{TwtrDetective} predicts it as a an inconsistent pair and points out speech is manipulated since $c^{vs}$ and $c^{cs}$ are smaller. }
    \vspace{-0.2in}
    \label{fig:model}
\end{figure*}

\noindent\textbf{Inconsistent Speech Generation.} Unlike \cite{tan2020detecting} and \cite{fung2021infosurgeon} editing multiple parts of the article, our target is to only modify the evidence sentences from the speech to reduce the linguistic bias. First, we use cosine similarity on SBERT sentence embeddings to extract the most similar sentences to the real claims. After that, we manipulate the named entities or event elements within the evidence sentences.

\noindent\textbf{Inconsistent Video Generation.} As for the third type of inconsistent pairs, we keep the speech and claims as the original ones. However, we replace the original videos with another real video that is similar to the current one by using the adversarial matching \cite{liu2020violin} method. Specifically, we utilize pretrained VideoCLIP \cite{xu2021videoclip}  to generate video representations to calculate the cosine similarity.

\noindent\textbf{Missing Modalities. } In order to handle the video-claim pairs missing the speech text, we select the real speech text from the dataset which has common entities with respect to the current claim. This new tuple with the real video, fake speech and real claim will be regarded as the inconsistent pairs in \texttt{COVID-VTS}. Examples generated by our tool are shown in Figure~\ref{fig:Automatic_tool} and Figure~\ref{fig:visual1}. Without the in-depth reasoning and cross-modal understanding, it's challenging to distinguish them from the real pairs. 

After all the filtering steps of automatic quality and bias control with manual validation, we assign the same amount of manipulated samples with pristine ones. Thus, the resulted dataset consists of 10k well-formed claims with associated videos.

\subsection{Dataset Analysis}
\texttt{COVID-VTS} exhibits three important differences to current benchmark datasets for video-text tasks \cite{xu2016msr, chen2011collecting, rohrbach2015dataset, anne2017localizing, krishna2017dense}. First, \texttt{COVID-VTS} brings new challenges with more diverse and complex scenes, including indoor press conference, news broadcasting, outdoor activities like protests, interviews, and screen recordings as well a slide shows. In contrast to recent video-text datasets \cite{xu2016msr, chen2011collecting, rohrbach2015dataset, anne2017localizing, krishna2017dense}, \texttt{COVID-VTS} has more available speech and screen text in Figure~\ref{fig:available}. Specifically, 87.5\% of the videos have speech and over 77\% of the videos have the screen text.  \texttt{COVID-VTS} has more named entities and videos (Table~\ref{tab:comparison_with_VIOLIN}). 90.8\% of the sentences in \texttt{COVID-VTS} have named entities while VIOLIN \cite{liu2020violin} is 10.2\%. This large gap indicates \texttt{COVID-VTS} not only is an excellent resource to support   research in the alignment between videos and named entities, but also provides new challenges to existing video-language inference models. Finally, \texttt{COVID-VTS} is the first fact-checking benchmark on the short video platform with inconsistent information from different modalities, which makes the fact-checking task more challenging.

\label{sec:dataset}

\section{Fact-Checking Model}
\subsection{Overview}
As shown in Figure~\ref{fig:model}, our fine-grained fact-checking system, \texttt{TwtrDetective}, is able to evaluate the overall factual consistency by integrating pairwise relation embeddings between different modalities. \texttt{TwtrDetective} also points out which modality is inconsistent and provides explanations to support the verification.

\noindent\textbf{Feature Extractor}. We use the vision transformer of CLIP (ViT-B/32) \cite{radford2021learning} to encode every frame into features. In particular, it extracts $N$ non-overlapping image patches from the frame and perform linear projection to map every patch into 1D token $\{v^p_1, \dots, v^p_N\}$, where $v_i\in\mathbb{R}^{D}$, where $D=512$, $N$ is the number of patches for each frame.  Second, we apply pretrained RoBERTa \cite{liu2019roberta} to generate contextual text representations, which utilizes byte pair encoding \cite{shibata1999byte} to tokenize the sentences. Therefore, each sentence in the speech content $S$ is represented as a sequence of tokens $\{t^s_1, \dots, t^s_K\}$, where $t^s_i\in\mathbb{R}^{D}$ and $D=768$ and screen text $O$ are $\{t^r_1, \dots, t^r_M\}$, where $t^r_i\in\mathbb{R}^{D}$ and $D=768$. Finally tokens in the claim $C$ are defined as $\{t^c_1, \dots, t^c_I\}$, where $t^c_i\in\mathbb{R}^{D}$ and $D=768$.

\noindent\textbf{Event Alert Module} The intuitive method to check the consistency is to calculate the video-text similarity with video-sentence retrieval models \cite{liu2019use, gabeur2020multi}. However, the  performance is dismal if the true information is tampered by small fragments. This is because existing models mainly encode the text at the document level without giving sufficient signal to focus on the factual elements, namely the event trigger and event argument. 

To provide explicit guidance to learn the internal event semantic of the text, we first employ DYGIE++ \cite{hope2020extracting} to detect the event structures within the text, assigning 1 if the token is the event trigger, 2 if it's the event argument and 0 otherwise. The indices are then fed into the learnable embedding table to generate \textit{Event Alert} gate $g^{alert}$, where $g^{alert}\in\mathbb{R}^{D}$ and $D=512$. A key property of $g^{alert}$ is that it helps our model determine the salience of tokens in the text.
Then, the claim token $t^c_i$, speech token $t^s_i$, OCR token $t^r_i$ are projected into a common semantic subspace with the same dimension by learning parameters $W^{c}$, $W^{s}$, $W^{r}$.
\begin{align}
{t^{c'}_i} &= g^{alert} \odot \text{tanh}(W^{c}{t^c_i})\\
{t^{s'}_i} &= g^{alert} \odot \text{tanh}(W^{s}{t^s_i})\\
{t^{r'}_i} &= g^{alert} \odot \text{tanh}(W^{r}{t^r_i})
\end{align}
where $\odot$ represents the element-wise multiplication operation and $tanh$ is the activation function. 

\noindent\textbf{Multi-head AoA Layer}. Motivated by the architecture presented in \cite{liu2020visual}, we contextualises $t^{c'}_i$, $t^{s'}_i$, $t^{r'}_i$ by using stacked Multi-Head Attention on Attention Layer, which takes advantage of the "Attention on Attention" module \cite{huang2019attention} to facilitate the generation of attended information. After encoding the text, the output of \textsc{[cls]} tokens named as $s^c$, $s^s$ and $s^r$ are utilized as the sentence representations of claim, speech content and screen text correspondingly.
\begin{align}
s^c &= \text{MHAoA}^\text{Mask}(\{t^{c'}_i, \dots, t^{c'}_I\})_{[cls]}\\
s^s &= \text{MHAoA}^\text{Mask}(\{t^{s'}_i, \dots, t^{s'}_K\})_{[cls]}\\
s^r &= \text{MHAoA}^\text{Mask}(\{t^{r'}_i, \dots, t^{r'}_M\})_{[cls]}
\end{align}
Patch features are also projected into the common subspace with the text by $W^{v}$. In order to learn the salience of patches in each frame, we feed the patch features with injection of positional information into Multi-Head AoA Layer to model the correlation of each patch. After that, we leverage the global average pooling to output the representation of each frame $v^f$.
\begin{align}
v'_i &=  \text{tanh}(W^{v}v_i)\\
v^f &= {Pool}(\text{MHAoA}^\text{Mask}(\{v'_1, \dots, v'_N\}))
\end{align}

\noindent\textbf{Pairwise Consistency Aggregation (PCA)}. To model the consistency between the claim, speech and video frames, we apply the Cross-modal Transformer \cite{li2020hero} learn the pairwise relationship. First, we fed the speech sentence embeddings $S=\{s^s_1 \dots, s^s_{L_s}\}$, visual frame embeddings $F=\{ v^f_1 \dots, v^f_{L_f}\}$ and its associated OCR sentence embeddings $O=\{s^r_1 \dots, s^r_{L_r}\}$ as the input. $L_s$, $L_f$, $L_r$ present the number of sentences. We also add the \textsc{[cls]} token in the first place of the input sequence. The outputs from Cross-modal Transformer is a sequence of contextualized embeddings. We use the output from the \textsc{[cls]} token represent the consistency relationship between the video and speech.
\begin{align}
R^{vs}&= \text{Cross-Transformer}(F, O, S)\\
R^{vc}&= \text{Cross-Transformer}(F, O, C)\\
R^{cs}&= \text{Cross-Transformer}(C, S)
\end{align}
As for the consistency measurement between the claim $C=\{s^c_1 \dots, s^c_{L_c}\}$ and speech $S$, video $F$, we also utilize the Cross-modal Transformer to integrate different modalities and \textsc{[cls]} to represent the weight. The claim can be regarded as the pointer to precisely retrieve relevant evidences from the paired video and speech, which play a key role to infer the truthfulness of the pairs.

\noindent\textbf{Explanations}.
After feeding pairwise consistency embeddings $R^{vs}$, $R^{vc}$ and $R^{cs}$ into MLP layers and sigmoid functions, our model produces the relation scores $c^{vs}$ which represents the consistency score between video and speech, $c^{vc}$ which represents the consistency score between video and claim and $c^{cs}$, which represents the consistency score between claim and speech. If a video post is classified as \texttt{inconsistent} by our model, the common modality of the two links with lower scores will be pointed as the \texttt{inconsistent} modality. For example in Figure~\ref{fig:model}, \texttt{TwtrDetective} detects that speech is inconsistent since $c^{vs}$ and $c^{cs}$ are smaller. It can be used the explanation to support the judgement.

\noindent\textbf{Objective Function}.
We optimize our model by minimizing the standard cross-entropy as shown on the top of Figure~\ref{fig:model}, where $y$ is the ground truth label and $y^P$ is the prediction probability after putting $c^{vs}$, $c^{vc}$ and $c^{cs}$ into the transformer attention layer.

\label{sec:model}

\section{Experiments}
\label{sec:experiment}

\begin{table}[t]
\setlength\tabcolsep{4pt}
\centering
\small
\begin{tabular}{lccc}
\toprule
Model & Accuracy & F1\\
\midrule
CLIP4Clip \cite{luo2021clip4clip} & 57.3 & 55.8\\
CLIP2Video \cite{fang2021clip2video} & 59.4 & 56.8\\
VideoClip \cite{xu2021videoclip} & 50.4 & 49.5\\
McCrae \cite{mccrae2022multi} & 62.7 & 62.3\\
MTS \cite{liu2020violin} & 56.3 & 55.4 \\
MMT \cite{gabeur2020multi} & 61.2 & 60.6\\
\bf{TwtrDetective (Ours)} & \bf{68.1} & \bf{67.9}\\
\midrule
Human Check & 82.7  & 81.9 \\
\bottomrule
\end{tabular}
\caption{
Comparative results (\%)  with baselines on the COVID-VTS Dataset.
}
\vspace{-0.1in}
\label{tab:COVID-VTS-result}
\end{table}

\begin{table}[t]
\setlength\tabcolsep{11pt}
\centering
\small
\begin{tabular}{lc}
\toprule
Model & Accuracy\\
\midrule
MTS \cite{liu2020violin}        & 60.4\\
XML \cite{lei2020tvr}          & 69.6 \\
HERO \cite{li2020hero}         & 70.4\\
\bf{TwtrDetective (Ours)}         & \bf{72.6}\\

\bottomrule
\end{tabular}
\caption{
Comparative results (\%) with baselines on the VIOLIN Dataset.
}
\label{tab:VIOLIN-result}
\end{table}

\begin{table}[t]
\setlength\tabcolsep{1.5pt}
\centering
\small
\begin{tabular}{lcc}
\toprule
Model & Accuracy & F1\\
\midrule
TwtrDetective &  68.1 & 67.9 \\
TwtrDetective (w/o Event Alert Module) & 64.4 & 64.5\\
TwtrDetective (w/o PCA) & 66.3 & 65.9\\
\bottomrule
\end{tabular}
\caption{
Ablation study to investigate our model's performance without the event alert module or the pairwise consistency aggregation module.
}
\vspace{-0.2in}
\label{tab:modules}
\end{table}

\begin{figure*}[t]
    \centering
      \includegraphics[width=0.96\textwidth]{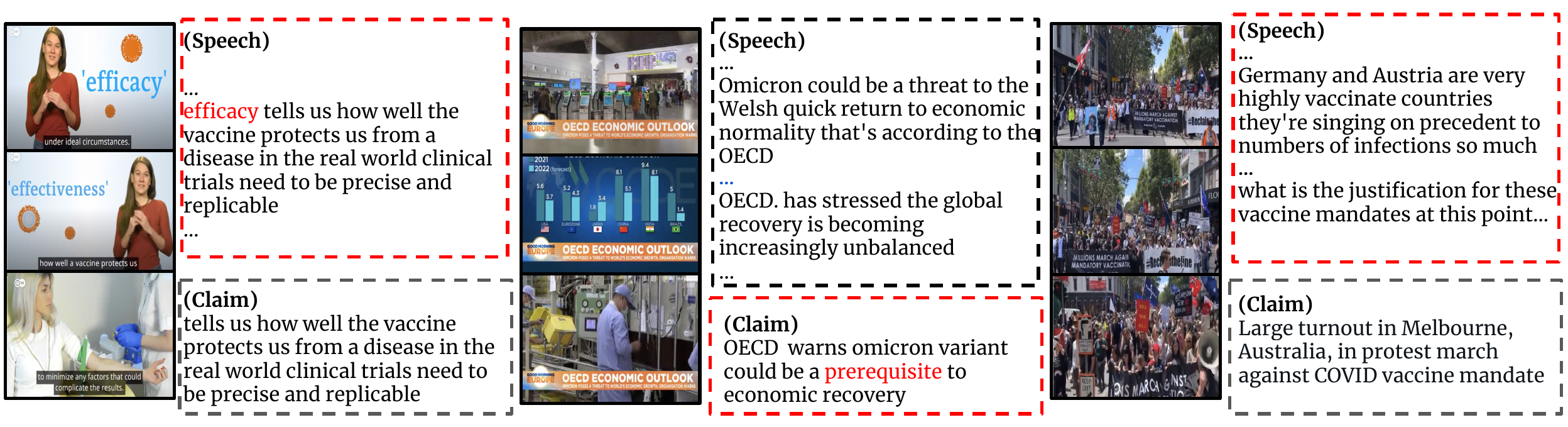}
    \caption{Qualitative analysis results. The red box indicates the inconsistent modality. \texttt{TwtrDetective} predicts the first two examples correctly except the third one.}
    \vspace{-0.1in}
    \label{fig:visual1}
\end{figure*}

In this section, we first introduce details of our implementation and compare the results to competing methods. Lastly, we present comprehensive experiment results and discussions. 

\subsection{Implementation Details}
\noindent\textbf{Datasets.}~We conduct experiments on two datasets: (1) COVID-VTS dataset,  which consists of 10k video-text pairs with different taxonomies generated by our automatic tool. Half of the samples are pristine and half are manipulated. (2) VIOLIN dataset \cite{liu2020violin}, a Video-and-Language Inference, collected from TV shows and movies, which contains 15,887 video clips and claims written by human. We also generate inconsistent speech and video pairs with our automatic tool so as to make sure the inconsistency comes from different modalities. Our model is compared with baselines on both datasets.

\noindent\textbf{Model Training.}~Our model is implemented using Pytorch. In our implementation, the dimensions for the common subspace is 512. Models are optimized using Adam with learning rate as $0.0005$. In addition,  we adopt a uniform sampling strategy to extract the frames and the sampling rate is 1 frame per second. In all the experiments, we split the COVID-VTS dataset into 80\% for training,  10\% for validation and 10\% for testing.

\noindent\textbf{Evaluation Metric.}~We compute the accuracy and F1 score at distinguishing inconsistent pairs from consistent ones.

\noindent\textbf{Baselines}. (1) CLIP4Clip~\cite{luo2021clip4clip} (2) CLIP2Video~\cite{fang2021clip2video} uses CLIP \cite{radford2021learning} to encode visual frames and text and use the transformer to fuse temporal information. (3) VideoClip~\cite{xu2021videoclip} is a state-of-art zero-shot video and text understanding model, which learns fine-grained association between video and text in a transformer. (4) McCrae ~\cite{mccrae2022multi} employs Long LSTM to integrate the video, text and named entity information from the video posts. (5) MMT, a video-text model that designs a multi-modal transformer to jointly encode the different modalities in video. (7) HERO \cite{li2020hero}: a transformer-based framework with two standard hierarchies for local and global context computation.

\subsection{Results and Discussion}
\noindent\textbf{Comparison Experiments.}~As Table~\ref{tab:COVID-VTS-result} and Table~\ref{tab:VIOLIN-result} show, our model achieves SOTA results compared with baselines on the both \texttt{COVID-VTS} and VIOLIN dataset. CLIP-variant models fail to perform well, revealing they do not detect token-level manipulation by analyzing the semantics of the sentence level. We notice that McCrae and MMT outperform MTS on both scores. This is because McCrae and MMT extract additional features from videos to help verification like objective detection and face detection. However, named entity verification proposed by McCrae does not show improvement on our since because it's incapable of finding the event-trigger manipulation. The main advantages of our model is that our event alert module is able to extract fine-grained factual details from the heterogeneous content. The pairwise consistency aggregation module is able to measure the interaction between each modality precisely. These observations also explain  why our model outperforms HERO, which ignores the guide from the event structures. The performance on the VIOLIN dataset is better compared to the results on \texttt{COVID-VTS} dataset. This is because our \texttt{COVID-VTS dataset} is more challenging with more named entities and events.

\begin{table}[t]
\setlength\tabcolsep{3pt}
\centering
\small
\begin{tabular}{lccc}
\toprule
Model & Accuracy & F1\\
\midrule
TwtrDetective  & 68.1 & 67.9\\
TwtrDetective ( w/o $Pair_{[Claim,Video]}$ ) & 66.6 & 66.3\\
TwtrDetective ( w/o $Pair_{[Claim,Speech]}$ )  & 67.2 & 67.0\\
TwtrDetective ( w/o $Pair_{[Speech,Video]}$ )  & 67.5 & 67.2\\
\bottomrule
\end{tabular}
\caption{
Analysis on the importance of different modality pairs in the PCA module.
}
\vspace{-0.2in}
\label{tab:modality}
\end{table}

\noindent\textbf{Analysis on Different Modalities.} We gain further insights into the importance of different modality pairs in PCA module. Our model in the first row of Table~\ref{tab:modality} uses all three pairs $Pair_{[Claim,Video]}$, $Pair_{[Claim,Speech]}$ and $Pair_{[Speech,Video]}$ in PCA. Other rows miss one of these pairs. Table~\ref{tab:modality}  indicates the PCA module is effective to improve the consistency checking performance between multiple modalities. In addition, $Pair_{[Claim,Video]}$ contributes  more than the other two pairs. This is because the evidence sentences are the only manipulated parts in the long speech text, making it challenging to be detected by referring to claims and videos.

\begin{figure}[t]
    \centering
      \includegraphics[width=0.4\textwidth]{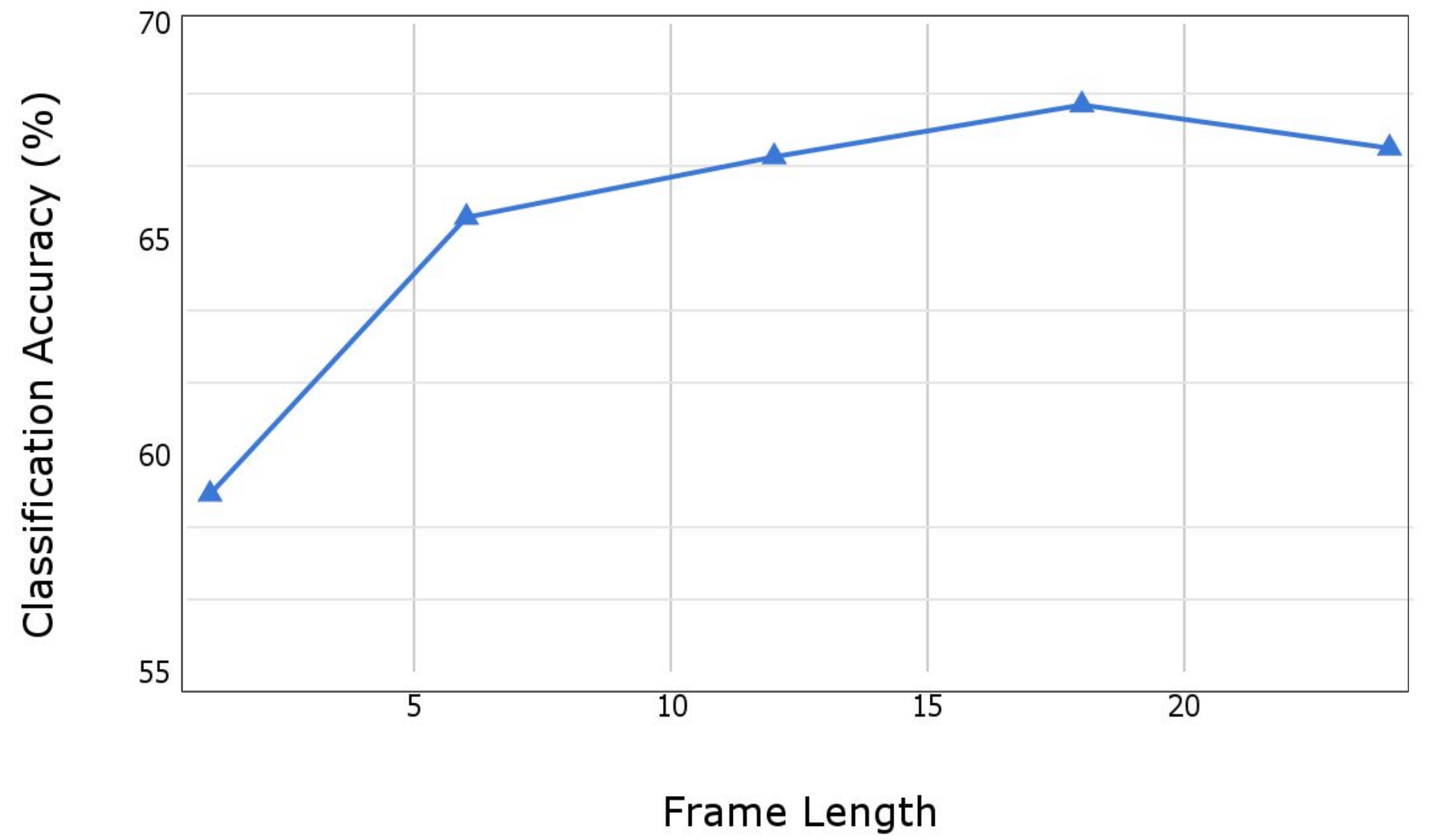}
    \vspace{-3pt}
    \caption{Ablation study to investigate our model's performance with different frame length.
    }
    \label{fig:length}
\end{figure}

\noindent\textbf{Ablation Study}. We investigate frame length in Figure~\ref{fig:length}, we can see a significant increase between 1 and 6 frames which shows it is better to encode the videos with a sequence of multiple frames instead of one single frame. We sampled 18 frames for our experiment, which is both efficient and effective. Furthermore in Table~\ref{tab:modules}, our model mainly benefits from \textit{Event Alert} module which provides 3.7\% boost in classification accuracy by explicitly tracking distorted factual pieces at the token level, \textit{PCA} module contributes 1.8\% improvement compared to using one-stream transformer to aggregate features from different modalities. Additionally, our model achieves better accuracy on short claims (length < 15) than on long claims (length > 25). This is because long claims have more event triggers and arguments than short claims, which makes it challenging for our event alert module to capture the manipulated tokens. 

\noindent\textbf{Qualitative Analysis.}
Figure~\ref{fig:visual1} presents prediction examples from our model. The correct cases (first two examples) demonstrate the model's ability to recognize the tiny inconsistent tokens. Since the original pair of third example miss the speech modality, we add the speech text which shares more common entities with the claim as the inconsistent modality. \texttt{TwtrDetective} fails to predict third post, suggesting that it does not work well on the video and speech alignment. The reasons behind might because both the video and speech describe the protest against the vaccine mandate but in different countries. Only from the video, it’s hard to detect whether it’s in Australia, Germany or Austria. In addition, the date information is also challenging to check.  The future direction could be asking the model to point out which part of the information is inconsistent or unverifiable and define the inconsistency taxonomy for them.

\section{Conclusions}
\vspace{-0.05in}
In this paper, we release a new benchmark, \texttt{COVID-VTS}, for fact-checking in short video platforms, consisting of 10k video-claim pairs. We develop an efficient tool to automatically generate large-scale trustworthy inconsistent pairs with different semantic meanings. Furthermore, our proposed fact-checking model \texttt{TwtrDetective} achieves state-of-the-art detection accuracy. We hope this work paves the way for future studies in multi-modal fact-checking as well as other related research areas in video and language.

\section{Ethical Statement}
Our goal in developing state-of-art consistency checking technique is to enhance the field’s ability to detect fake news and improve the Twitter community health. According to Twitter Developer Policy, Twitter supports the research that measures and analyzes topics like spam, abuse, or other platform health-related topics for non-commercial research purposes. In addition, the posts we collected are from verified and authentic accounts. Certain accounts are ineligible for the verified badge if they post content that harasses, shames, or insults any individual or group, or violate the Platform manipulation and spam policy. To protect the personal information, we will only use the captions and videos as input without the user information. We also work to filter the dataset and only keep the posts discussing public news instead of personal life. Personal information but not limited to, user’s name, health, financial status, racial or ethnic origin, religious or philosophical affiliation or beliefs, sexual orientation, trade union membership, alleged or actual commission of crime. It’s crucial to mention that we will not share our source video file but the URL links or the extracted features from the videos. This is to due to the licence policy and avoid anyone to deliberately generate and spread misinformation. As such, we will release the model code but not the output in our work for public verification and auditing so it can be used to combat fake news.
\label{sec:conclusion}

\section{Limitations}
There is a significant gap between our model and human performance on the accuracy. We hope \texttt{COVID-VTS} dataset will encourage the community to develop stronger models in the future. One possible direction is to develop models to localize key frames or key sentences from the speech to deduce the difficulty of consistency check.

\label{sec:conclusion}

\section*{Acknowledgements}
This work was supported by DARPA SemaFor (HR001119S0085) program.
We are thankful for the feedback from anonymous reviewers of this paper.
\label{sec:Acknowledgements}

\bibliography{anthology,custom}
\bibliographystyle{acl_natbib}

\end{document}